\documentclass[runningheads]{llncs}
\usepackage{subcaption}
\usepackage{graphicx}
\usepackage{comment}
\usepackage{amsmath,amssymb} 
\usepackage{xcolor}
\usepackage{mathtools}
\usepackage{booktabs}
\usepackage[export]{adjustbox}
\usepackage{hyperref}

\newcommand{\ci}[1]{$\scriptstyle{\pm #1}$}

\newcommand{\img}{\mathbf{x}}
\newcommand{\ds}{\mathcal{D}}
\newcommand{\feat}{\mathbf{z}}
\newcommand{\reals}{\mathbb{R}}

\newcommand{\pfeat}{\widetilde{\feat}}

\begin{document}
\pagestyle{headings}
\mainmatter
\def\ECCVSubNumber{5313}  

\title{Embedding Propagation: Smoother Manifold for Few-Shot Classification} 

\titlerunning{Embedding Propagation}
%
\author{Pau Rodr\'{i}guez\inst{1} \and
Issam Laradji\inst{1,2} \and
Alexandre Drouin\inst{1} \and
Alexandre Lacoste\inst{1}
}
\authorrunning{P. Rodríguez et al.}
%
\institute{Element AI, Montreal, Canada \and
University of British Columbia, Canada\\
\email{\{pau.rodriguez,issam.laradji,adrouin,allac\}@elementai.com}}
\maketitle

\begin{abstract}
Few-shot classification is challenging because the data distribution of the training set can be widely different to the test set as their classes are disjoint. This distribution shift often results in poor generalization. Manifold smoothing has been shown to address the distribution shift problem by extending the decision boundaries and reducing the noise of the class representations. Moreover, manifold smoothness is a key factor for semi-supervised learning and transductive learning algorithms. In this work, we propose to use embedding propagation as an unsupervised non-parametric regularizer for manifold smoothing in few-shot classification. Embedding propagation leverages interpolations between the extracted features of a neural network based on a similarity graph. We empirically show that embedding propagation yields a smoother embedding manifold. We also show that applying embedding propagation to a transductive classifier achieves new state-of-the-art results in \textit{mini}Imagenet, \textit{tiered}Imagenet, Imagenet-FS, and CUB. Furthermore, we show that embedding propagation consistently improves the accuracy of the models in multiple semi-supervised learning scenarios by up to 16\% points. The proposed embedding propagation operation can be easily integrated as a non-parametric layer into a neural network. We provide the training code and usage examples at~\url{https://github.com/ElementAI/embedding-propagation}.

\end{abstract}

Deep learning methods have achieved state-of-the-art performance in computer vision tasks such as classification \cite{krizhevsky2012Imagenet}, semantic segmentation~\cite{long2015fully}, and object detection~\cite{ren2015faster}. However, these methods often need to be trained on a large amount of labeled data. Unfortunately, labeled data is scarce and its collection is expensive for most applications. This has led to the emergence of deep learning methods based on transfer learning~\cite{yosinski2014transferable}, few-shot learning (FSL)~\cite{fei2006one}, and semi-supervised learning~\cite{chapelle2005semi}, that address the challenges of learning with limited data.

\begin{figure}
    \centering
    \includegraphics[width=0.8\linewidth]{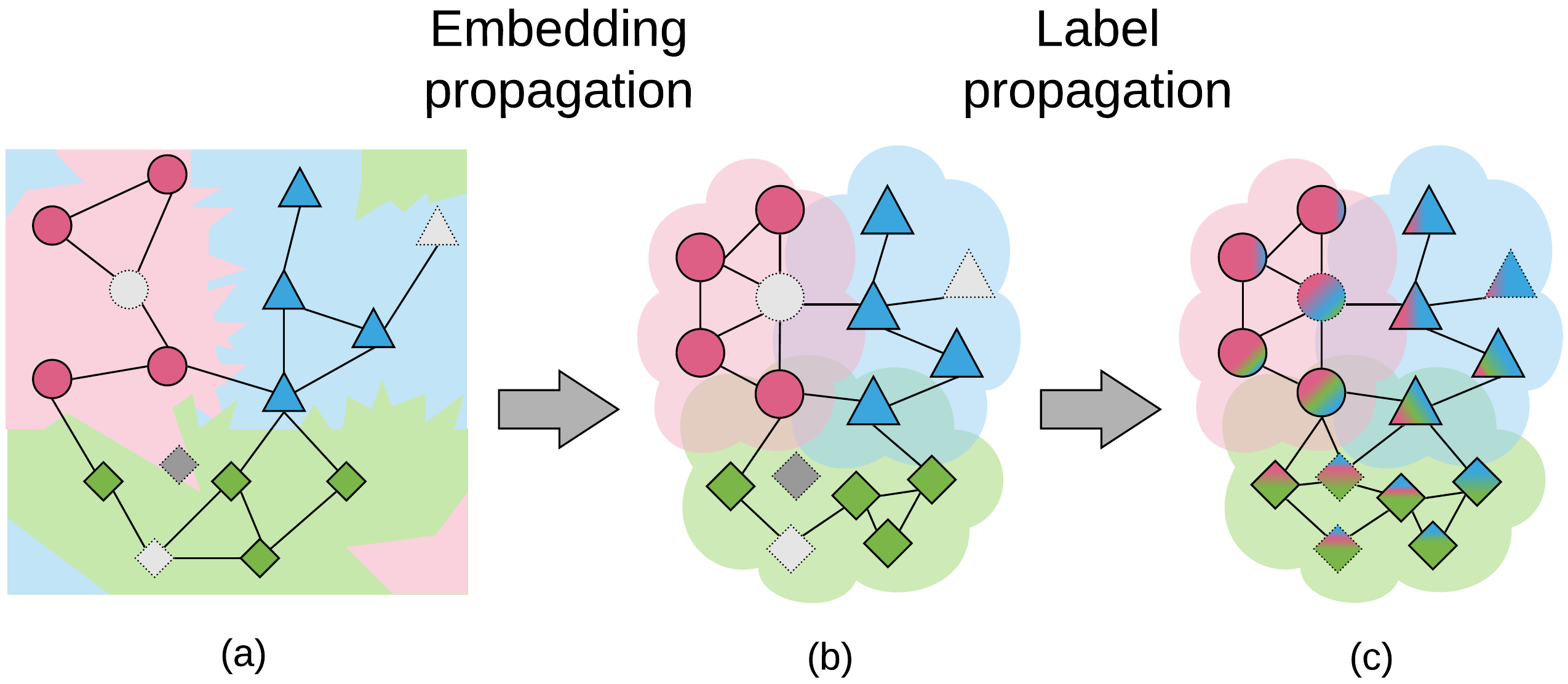}
    \caption{Illustration of the embedding propagation (EP) method. (a) Original decision boundaries for three classes. The color of a region represents the predicted class, and the color of a node represents the node's actual class. (b) Decision boundaries after applying EP, which are smoother than in (a). (c) Predictions after propagating the labels across the graph, leveraging unlabeled points (light gray) to classify a query example (shown in dark gray). \textit{Best viewed in color}.}
    \label{fig:embeddingprop-sketch}
    
\end{figure}

Few-shot learning methods have the potential to significantly reduce the need for human annotation. This is because such methods learn new tasks with few labeled examples by transferring the knowledge gained across several tasks.
Three recent approaches have been successful for few-shot classification (FSC): metric learning, meta learning, and transfer learning. Metric learning approaches~\cite{vinyals2016matching,snell2017prototypical} learn an embedding space where a set of labeled examples (\textit{support set}) is used to predict the classes for unlabeled examples (\textit{query set}). Meta-learning approaches~\cite{finn2017model,rusu2018meta} learn to infer a set of parameters that can be adapted to new tasks. Transfer learning~\cite{chen2018a,mangla2019charting} methods aim to learn a general feature representation and then train a classifier for each new task. In this work, we use an approach between metric learning and transfer learning. During training, the model attempts to learn a general feature representation that is fine-tuned using a metric-based classifier.

A key challenge in few-shot classification is training models that generalize well to unseen classes. This requires feature representations that are robust to small changes in the data distribution. This issue has been addressed outside the few-shot learning literature with a number of regularization techniques such as dropout \cite{srivastava2014dropout}, batch normalization \cite{ioffe2015batch}, and manifold mixup \cite{verma2019manifold}. However, regularization in few-shot learning remains unexplored. In this work, we show that re-framing label propagation to perform manifold smoothing improves the performance of few-shot classifiers, particularly in the transductive and semi-supervised settings. Different from manifold mixup \cite{verma2019manifold}, the proposed method is unsupervised and captures higher order interactions between the embedding.

We propose an embedding propagation (EP) method that outputs a set of interpolations from the network output features using their similarity in a graph. This graph is constructed with pairwise similarities of the features using the radial basis function (RBF). EP is non-parametric and can be applied on top of any feature extractor. It can be used as part of a network in order to obtain a regularized manifold for both training and testing. We refer to such network as EPNet. For few-shot classification, we empirically show that the proposed regularization improves the performance for transductive and semi-supervised learning. The hypothesis behind this improvement is based on the fact that using interpolated embeddings result in smoother decision boundaries and increased robustness to noise. These properties have been shown to be important for generalization~\cite{bartlett1999generalization,lee1995lower,verma2019manifold} and semi-supervised learning \cite{chapelle2005semi}.

For semi-supervised learning (SSL), EPNet takes advantage of an unlabeled set of images at test time in order to make better predictions of the query set. We adapt the SSL approach proposed by Lee \emph{et al.}~\cite{lee2013pseudo} to the few-shot classification setup. Thus, for each unlabeled image, EPNet selects the class that has the maximum predicted probability as the pseudo label. EPNet then uses these pseudo labels along with the support set to perform label propagation to predict the labels of the query set. This approach achieves significant improvement over previous state-of-the-art in the 1-shot SSL setting. We hypothesize that EPNet is effective in the SSL setting because of the properties of smoother manifolds~\cite{chapelle2005semi}. 

Overall, EPNet achieves state-of-the-art results on \textit{mini}Imagenet~\cite{vinyals2016matching},\\ \textit{tiered}Imagenet~\cite{ren2015faster}, Imagenet-FS~\cite{hariharan2017low} and CUB~\cite{WelinderEtal2010} for  few-shot classification, and semi-supervised learning scenarios. In our ablation experiments, we evaluate different variations of embedding propagation and their impact on the smoothness of the decision boundaries. We also show that, with EP, we also achieve a clear improvement on the SSL setup compared to the same model without EP. 


Our main contributions can be summarized as follows. We show that  embedding propagation:

\begin{itemize}
    \item Regularizes the manifold in an unsupervised manner.
    \item Leverages embedding interpolations to capture higher order feature interactions.
    \item Achieves state-of-the-art few-shot classification results for the transductive and semi-supervised learning setups.
\end{itemize}

\section{Related Work}
Our work focuses on few-shot classification, but also intersects with manifold regularization, transductive learning, and semi-supervised learning. We describe relevant work for each of these topics and point out their relevance to our method.

\paragraph{Few-shot classification.}
A common practice for training models for few-shot learning is to use  episodic learning \cite{ravi2016optimization,vinyals2016matching,snell2017prototypical}. This training methodology creates episodes that simulate the train and test scenarios of few-shot learning.

Meta-learning approaches make use of this episodic framework. They learn a base network capable of producing parameters for a task-specific network after observing the support set. The task-specific network is then evaluated on the query set and its gradient is used to update the base network. By doing so, the base network learns to use the support set to generate parameters that are suitable for good generalization. This was first introduced in \cite{ravi2016optimization}. Perhaps, the most popular meta-learning approach is MAML \cite{finn2017model} and other algorithms that derivate from it \cite{nichol2018reptile,rajeswaran2019meta},
which learn a set of initial weights that are adapted to a specific task in a small amount of gradient steps. However, this choice of architecture, while general, offers limited performance for few-shot image classification. This lead to variants of meta-learning methods more adapted to image classification \cite{oreshkin2018tadam,rusu2018meta,gidaris2018dynamic,qiao2018few}.

Most metric learning approaches are trained using episodes \cite{vinyals2016matching,snell2017prototypical,liu2018learning}, they can also be seen as meta-learning approaches. Concretely, metric learning approaches are characterized by a classifier learned over a feature space. They focus on learning high-quality and transferable features with a neural network common to all tasks. EPNet leverages the work of Liu \emph{et al.} \cite{liu2018learning} for learning to propagate labels, and thus falls into this category. Graph-based approaches can also be framed into this category~\cite{garcia2018fewshot,hu2020exploiting,yang2020dpgn,kye2020transductive}. Gidaris \textit{et al.}~\cite{gidaris2019generating} proposed to generate classification weights with a graph neural network (GNN) and apply a denoising autoencoder to regularize their representation. EP does also perform a regularization on a graph representation. Set-to-set functions have also been used for embedding adaptation~\cite{ye2020few}. However, different from GNNs and set-to-set, our graph is unsupervised and non-parametric, its purpose is manifold smoothing, and we show it improves semi-supervised learning approaches.

While metric learning offers a convenient approach to learn transferable features, it has been shown that neural networks trained with conventional supervised learning already learn transferable features \cite{bauer2017discriminative,chen2018a,mangla2019charting}. Hence, to learn a classifier on a new task, it suffices to fine-tune the feature extractor to that task. Also, this approach has shown to learn more discriminative features compared to the episodic scenario. To take advantage of this transfer learning procedure, we use it in our pre-training phase. Thus,  EPNet combines a metric-based classifier with the pre-training of transferable features to achieve a more general representation.

\paragraph{Regularization for Generalization.} 
Regularization is a principled approach for improving the generalization performance of deep networks. Commonly used techniques such as dropout \cite{srivastava2014dropout} and batch normalization \cite{ioffe2015batch} attempt to achieve robustness towards input variations. Others are based on regularizing weights \cite{rodriguez2016regularizing,salimans2016weight}. Another line of work that is based on manifold regularization~\cite{zhang2018mixup,verma2019manifold,tokozume2018between,belkin2006manifold}. These works propose methods that aim to smooth the decision boundaries and flatten the class representations, which are important factors for generalization~\cite{lee1995lower,bartlett1999generalization}. Similarly, we attempt to smooth the manifold by incorporating an embedding propagation operation on the extracted features during training. A concurrent work~\cite{mangla2019charting} and our work were the  first to apply manifold regularization on few-shot classification. However, the method presented in \cite{mangla2019charting} differs from ours in four ways. First, they perform smoothing in an additional training phase. Second, they train linear classifiers at inference time. Third, they use an exemplar self-supervised loss in their training procedure. Fourth, they do not show the efficacy of their method for semi-supervised learning. In the few-shot classification benchmarks, we achieve better classification accuracy on the Imagenet datasets and CUB dataset for the 1-shot, 5-shot, and 10-shot case.

 There are different lines of research showing that perturbing image representations results in better generalization \cite{srivastava2014dropout,ioffe2015batch}. The most closely related to our work are based on feature interpolation. For instance, Zhao and Cho \cite{cho2019retrieval} proposed to make predictions based on the interpolation of nearest neighbors to improve adversarial robustness. In \textit{Manifold Mixup} this idea was expanded to smooth the representations of the neural architecture and achieve better generalization \cite{bartlett1999generalization,lee1995lower}. \textit{Manifold Mixup} has been applied to FSC architectures as a  fine-tuning step to improve their performance \cite{mangla2019charting}. Differently, we propose a novel procedure to smooth the manifold end-to-end. The proposed method is applied only at the output layer and achieves higher classification accuracy than previous approaches. Moreover, and also different from \cite{mangla2019charting}, we leverage the properties of smoother manifolds for semi-supervised learning \cite{chapelle2009semi}, further widening the improvement margin.

\paragraph{Transductive learning (TL).}  The idea of transduction is to perform predictions only on the test points. In contrast, the goal of inductive learning is to output a prediction function defined on an entire space~\cite{vapnik1999overview}. Given a small set of labeled examples, transductive learning has been shown to outperform inductive learning~\cite{liu2018learning,iscen2019label,liu2019deep}. This makes TL a desirable method for few-shot classification. Liu \emph{et al.}~\cite{liu2018learning} presented one of the few work using TL for FSC. Similar to this work, we use label propagation~\cite{DBLP:conf/nips/ZhouBLWS03} to predict the labels of the query set. However, they do not incorporate a manifold smoothing method such as the embedding propagation method investigated in this work.

\begin{figure}[t!]
\centering
	\begin{subfigure}[t]{0.48\textwidth} 
	    \centering
		\includegraphics[width=\textwidth, left]{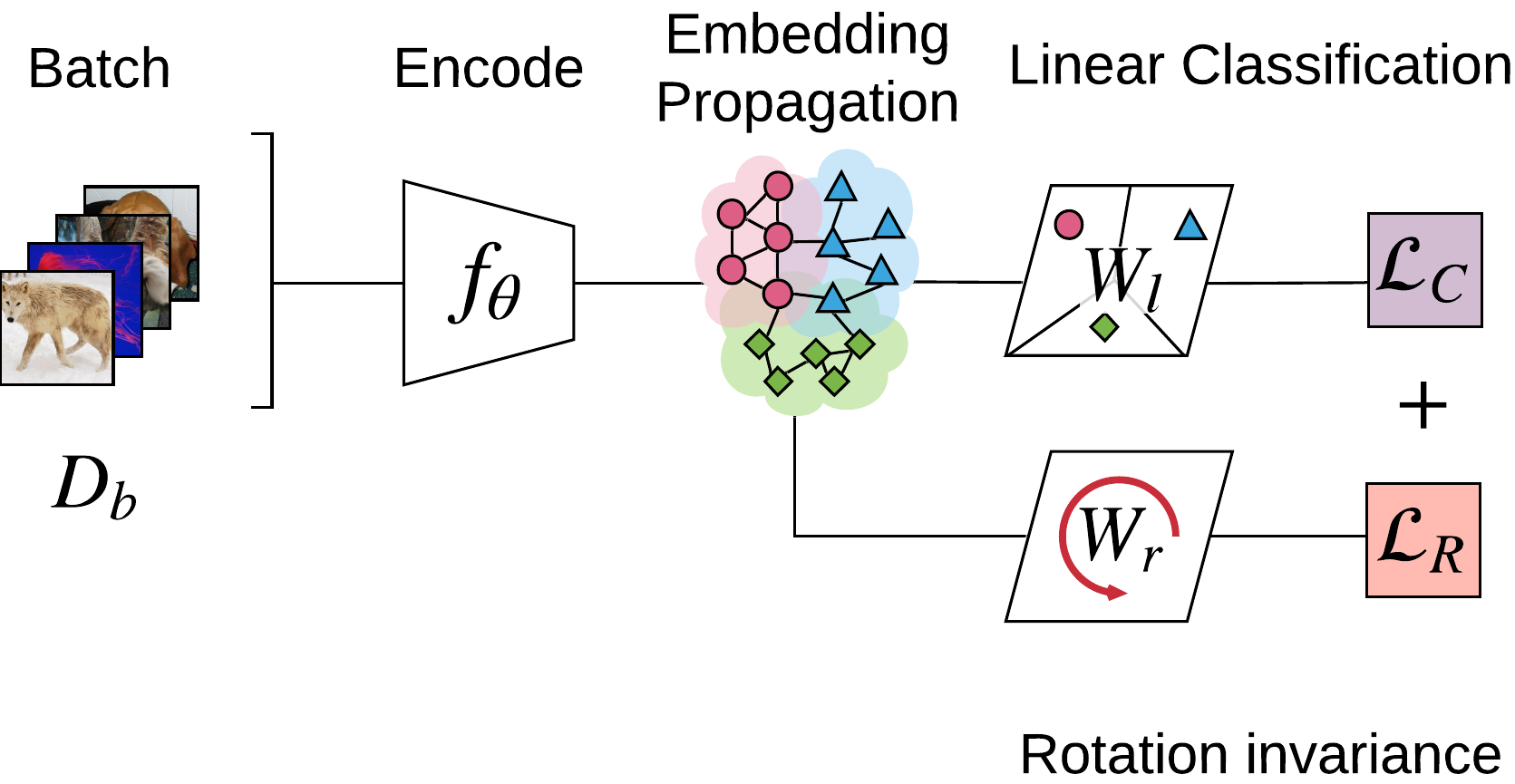}
		\caption{Pretraining phase} 
			\label{fig:method1}
	\end{subfigure}
	\hfill
	\begin{subfigure}[t]{0.48\textwidth} 
	    \centering
		\includegraphics[width=\textwidth, right]{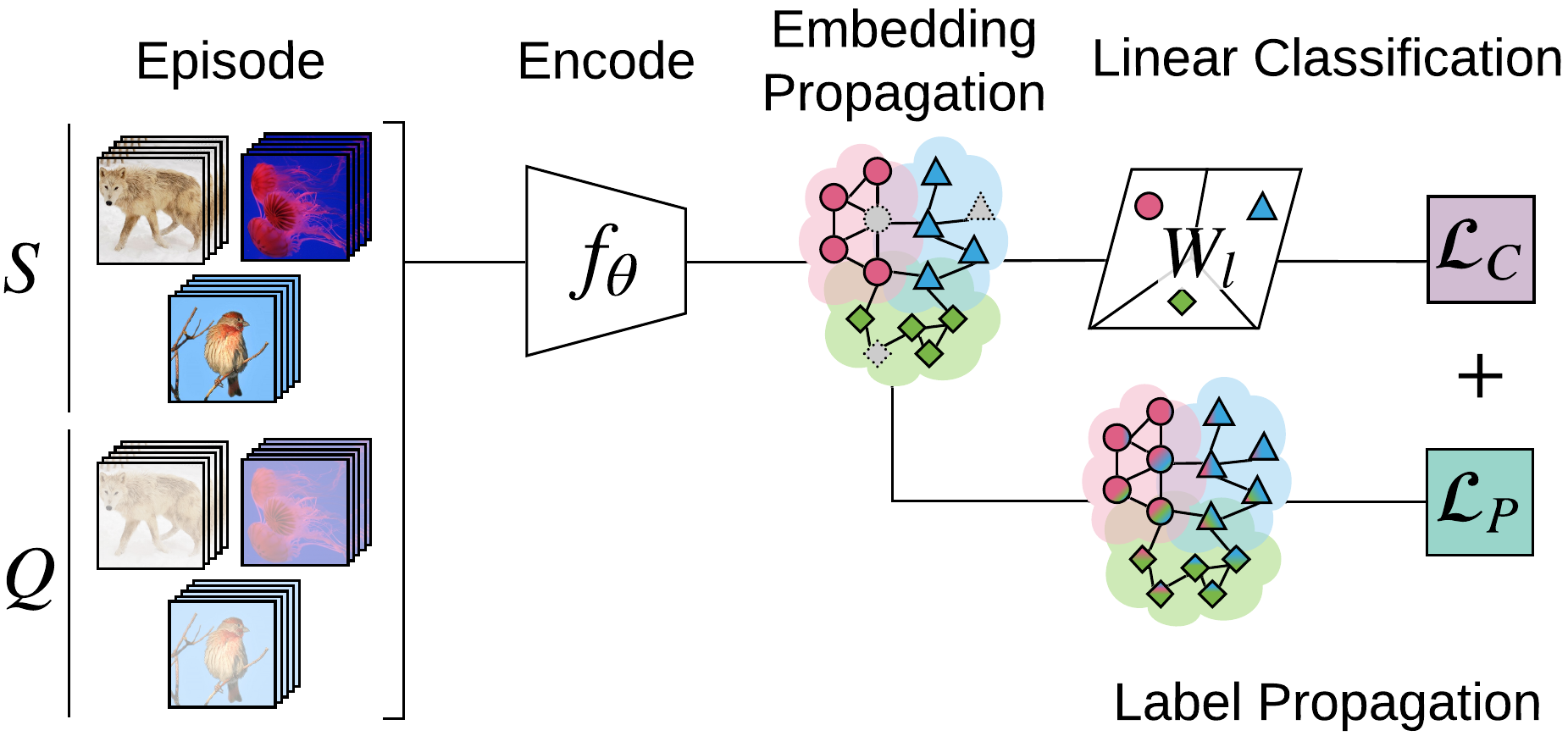}
		\caption{Episodic fine-tuning and evaluation} 
			\label{fig:method2}
	\end{subfigure}
	\caption{Overview of the EPNet training procedure. EPNet is trained in two phases: a pretraining phase and an episodic fine-tuning phase. \textbf{(a)} First, the model is trained to learn general feature representations using a standard classification loss $\mathcal{L}_C$ and an auxiliar rotation loss $\mathcal{L}_R$. \textbf{(b)} Then, the model is fine-tuned using episodic learning to learn to generalize to novel classes by minimizing the standard classification loss $\mathcal{L}_C$ and a label propagation loss $\mathcal{L}_P$. In both phases the features are encoded using a feature extractor followed by our proposed embedding propagation method (Sec.~\ref{sec:proposed_method}).} 
	\label{fig:method}
\end{figure}

\paragraph{Semi-Supervised learning.} 
While the literature on semi-supervised learning is vast, few works leverage the use of unlabeled data in few-shot image classification. In \cite{ren2018metalearning}, they develop a soft version of $k$-means to meta-learn how to use unlabeled data. Liu \emph{et al.} \cite{liu2018learning,liu2019deep} used label propagation to leverage a set of unlabeled data. Their approach works with both semi-supervised and transductive learning and improves results over soft $k$-means. In this work we use a similar label propagation approach and show that the semi-supervised results can be further improved by using pseudo-labels (labels obtained from the current model). Recently, Sun \emph{et al.} \cite{sun2019learning} used a meta-learning approach to cherry-pick examples from the unlabeled set and label them with the current model to increase the label set. In contrast, our method does not require cherry-picking which needs an extra learning step.
 
\section{Proposed Method}
\label{sec:proposed_method}
We propose an embedding propagation network (EPNet) that has the following pipeline. Given an input image, EPNet first extracts image features using a feature extractor. Then, we apply a novel embedding propagation method (described in Sec. \ref{sec:embedding-propagation}) to map the features to a set of  interpolated features that we refer to as embeddings. These embeddings are then used by a classifier to label the images (Sec. \ref{sec:inference}).  The goal of embedding propagation is  to increase the smoothness of the embedding manifold, which was shown to improve generalization \cite{bartlett1999generalization,lee1995lower} and the effectiveness of semi-supervised learning methods \cite{chapelle2009semi} (Sec. \ref{sec:method-semisup}). In the following sections, we explain EPNet in more detail.

\subsection{Embedding propagation} \label{sec:embedding-propagation}
Embedding propagation takes a set of feature vectors $\feat_i \in \reals^m$, obtained from applying a feature extractor (CNN) to the samples of an episode.  Then, it outputs a set of embeddings $\pfeat_i \in \reals^m$ through the following two steps. First, for each pair of features $(i, j)$, the model computes the distance as  $d^2_{ij} = \Vert \feat_i - \feat_j \Vert^2_2$ and the adjacency as $A_{ij} = \exp \left(-d^2_{ij}/\sigma^2\right)$, where $\sigma^2$ is a scaling factor and $A_{ii}=0, \; \forall i$,  as done in TPN \cite{liu2018learning}.  We chose $\sigma^2=  \operatorname{Var}\left(d^2_{ij}\right)$ which we found to stabilize training. 

Next we compute  the Laplacian of the adjacency matrix, 
\begin{equation}\label{eq:laplacian}
    L =D^{-\frac{1}{2}} A D^{-\frac{1}{2}},\;\; D_{ii} = \textstyle\sum_j A_{ij}.
\end{equation}

Finally, using the label propagation formula described in \cite{DBLP:conf/nips/ZhouBLWS03}, we obtain the propagator matrix $P$ as,
\begin{equation}
\label{eq:labelprop}
    P = (I - \alpha L)^{-1},
\end{equation}
where $\alpha \in \reals$ is a scaling factor, and $I$ is the identity matrix. Then, the embeddings are obtained as follows,
\begin{equation}
    \pfeat_i = \sum_j P_{ij} \feat_j.
    \label{eq:embeddings}
\end{equation}
Since the $\pfeat_i$ are now a weighted sum of their neighbors, embedding propagation has the effect of removing undesired noise from the feature vectors.
Note that this operation is simple to implement and compatible with a wide range of feature extractors and classifiers. Further, note that the computational complexity of Eq. \ref{eq:labelprop} is negligible for few-shot episodes \cite{liu2018learning} since the size of the episode is small.

\subsection{Few-shot classification setup}
Following the common few-shot setups \cite{vinyals2016matching,ren2018metalearning}, we are given three datasets: a \emph{base} dataset ($\ds_b$), a \emph{novel} dataset ($\ds_n$), and a \emph{validation} dataset ($\ds_v$). The base dataset is composed of a large amount of labeled images $\ds_b = \{(\img_i, y_i)\}_{i=1}^{N_{base}}$, where each image $\img_i$ is labeled with class $y_i \in \mathcal{Y}_{base}$. The novel dataset $\ds_n = \{(\img_j, y_j)\}_{j=1}^{N_{novel}}$, where $\img_j$ comes from previously unseen classes $y_j \in \mathcal{Y}_{novel}$, such that $\mathcal{Y}_{base} \cap \mathcal{Y}_{novel} = \emptyset$, is used to evaluate the transfer learning capabilities of the model. The validation dataset $\ds_{\operatorname{v}}$ contains classes not present in $\ds_b$ and $\ds_n$ and is used to conduct hyperparameter search. 

Furthermore, we have access to episodes. Each episode consists of $n$ classes sampled uniformly without replacement from the set of all classes, a support set $S$ ($k$ examples per class) and a query set $Q$ ($q$ examples per class). This is referred to as  $n$-way $k$-shot learning.

\subsection{Inference Phase}
\label{sec:inference}
Given an episode, we perform inference by extracting features of an input image, applying embedding propagation on those features,  then applying label propagation. 
More formally, this is performed as follows. Let $\widetilde{Z} \in \reals^{(k+q) \times m}$ be the matrix of propagated embeddings obtained by jointly applying Eq.\,\ref{eq:laplacian}-\ref{eq:embeddings} to the support and query sets.
Let $P_{\widetilde{Z}}$ be the corresponding propagator matrix.
Further, let $Y_S \in \mathbb{R}^{k \times n}$ be a one-hot encoding of the labels in the support set and $\mathbf{0} \in \mathbb{R}^{q \times n}$ a matrix of zeros. We compute the logits for the query set ($\hat{Y}_Q$) by performing label propagation as described in~\cite{DBLP:conf/nips/ZhouBLWS03}.

\subsection{Training procedure}
\label{sec:training-phase}
EPNet is trained in two phases as illustrated in Fig.~\ref{fig:method}. First, the model is trained on $\ds_b$ using the common pretraining procedure for few-shot classification~\cite{rusu2018meta} in order to learn a general feature representation. 
Second, the model is fine-tuned using episodes in order to learn to generalize to novel classes. Episodes are drawn from the same dataset $\ds_b$. In both phases, EPNet uses the same feature extractor $f_\theta(\img)$ parametrized by $\theta$ to obtain the features $\pfeat$ extracted for a given input image $\img$.
However, each phase relies on a different objective.


\paragraph{Pre-training phase.}
As shown in Fig. \ref{fig:method1}, we train $f_\theta$ using two linear classifiers, which are linear layers with softmax activations parametrized by $W_l$ and $W_r$, respectively. 
The first classifier is trained to predict the class labels of examples in $\ds_b$.
It is optimized by minimizing the cross-entropy loss,

\begin{equation}
\mathcal{L}_c(\img_i, y_i; W_l, \theta) = -\ln p(y_i|\pfeat_i, W_l),    
\end{equation}
where $y_i \in\mathcal{Y}_b$ and the probabilities are obtained by applying softmax to the logits provided by the neural network.

For fair comparison with recent literature, we also add a self-supervision loss \cite{mangla2019charting,gidaris2019boosting} to obtain more robust feature representations. Hence, we use the second classifier to predict image rotations and use the following loss,
\begin{equation}
\mathcal{L}_{r}(\img_i, r_j; W_r, \theta) = -\ln p(r_j|\pfeat_i, W_r),
\end{equation}
where $r_j \in \{0^\circ, 90^\circ, 180^\circ, 270^\circ \}$, and $p(r_j|\pfeat_i, W_r)$ is the probability of the input being rotated by $r_j$ as predicted by a softmax classifier with weights $W_r$.

Overall, we use stochastic gradient descent (SGD) with batches of size 128 and 4 rotations per image to optimize the following loss,
\begin{align}
\underset{\theta, W_l, W_r}{\operatorname{argmin}} \sum_{i=1}^{128} \sum_{j=1}^4 \mathcal{L}_c(\img_i, y_i; W_l, \theta) + \mathcal{L}_{r}(\img_i, r_j; W_r, \theta).
\end{align}
 
\paragraph{Episodic Learning phase.}
As shown in Fig. \ref{fig:method2}, after the pre-training phase, we use episodic training to learn to recognize new classes.
In this phase, we also optimize EPNet using two classifiers.
The first classifier is based on label propagation. It computes class probabilities by applying a softmax to the query set logits $\hat{Y}_Q$ defined in Sec. \ref{sec:inference}, i.e.,
\begin{equation}
\label{eq:lp-class}
\mathcal{L}_p(\img_i, y_i; \theta) = - \ln p(y_i| \pfeat_i, \widetilde{Z}, Y_S).
\end{equation}
The second classifier is identical to the $W_l$-based classifier used in pretraining.
It is included to preserve a discriminative feature representation.
Hence, we minimize the following loss:
\begin{align}
\underset{\theta, W_l}{\operatorname{argmin}} \left[ \tfrac{1}{\vert Q \vert} \sum_{\mathclap{(\img_i, y_i)\in Q}} \mathcal{L}_p(\img_i, y_i; \theta) + \tfrac{1}{\vert S \cup Q \vert} \sum_{\mathclap{(\img_i, y_i) \in S \cup Q}} \tfrac{1}{2}\mathcal{L}_c(\img_i, y_i; W_l, \theta) \right].
\end{align}


\subsection{Semi-supervised learning}\label{sec:method-semisup}
In the semi-supervised learning scenario, we also have access to an unlabeled set of images $U$. We use the unlabeled set as follows. First, we use the inference procedure described in Sec.~\ref{sec:inference} to predict the labels $\hat{c}_U$ for the unlabeled set as pseudo-labels. Then, we augment the support set with $U$ using their pseudo-labels as the true labels. Finally, we use the inference procedure in Sec.~\ref{sec:inference} on the new support set to predict the labels for the query set.

We also consider the semi-supervised scenario proposed by Garcia and Bruna \cite{garcia2017few}. In this scenario the model is trained to perform 5-shot 5-way classification but only 20\% to 60\% of the support set is labeled.

As shown by Lee \emph{et al.} \cite{lee2013pseudo}, this procedure is equivalent to entropy regularization, an effective method for semi-supervised learning. Entropy regularization is particularly effective in cases where the decision boundary lies in low-density regions. With embedding propagation we achieve a similar decision boundary by smoothing the manifold.

\section{Experiments}
\label{sec:experimental_setup}
In this section, we present the results on three standard FSC datasets, \textit{mini}Imagenet \cite{vinyals2016matching}, \textit{tiered}Imagenet \cite{ren2018metalearning}, CUB \cite{WelinderEtal2010}, and Imagenet-FS \cite{hariharan2017low}. We also provide ablation experiments to illustrate the properties of embedding propagation. As common procedure, we averaged accuracies on $\ds_n$ over 1000 episodes \cite{vinyals2016matching}.   

\subsection{Datasets}
\noindent\textbf{\textit{mini}Imagenet~\cite{ravi2016optimization}.} A subset of the Imagenet dataset \cite{ILSVRC15} consisting of 100 classes with 600 images per class. Classes are divided in three disjoint sets of 64 base classes, 16 for validation and 20 novel classes. \smallskip\\ 
\noindent\textbf{\textit{tiered}Imagenet~\cite{ren2018metalearning}.} A more challenging subset of the Imagenet dataset \cite{ILSVRC15} where class subsets are chosen from supersets of the wordnet hierarchy. The top hierarchy has 34 super-classes, which are divided into 20 base (351 classes), 6 validation (97 classes) and 8 novel (160 classes) categories.\smallskip\\
\noindent\textbf{Imagenet-FS \cite{hariharan2017low}.} A large-scale version of ImageNet. It is split into 389 base classes and 611 novel classes. The training set consists of 193 of the base classes. Validation consists of 193 of the base classes plus 300 novel classes. The test set consists of the remaining 196 base classes and the remaining 311 novel classes.\smallskip \\
\noindent\textbf{CUB \cite{chen2018a,hilliard2018few}.} A fine-grained dataset based on CUB200 \cite{WelinderEtal2010} composed of 200 classes and 11,788 images split in 100 base, 50 validation, and 50 novel classes.


\begin{table}[t!]
\centering
\caption{Comparison of test accuracy against state-of-the art methods for 1-shot and 5-shot classification using \textit{mini}Imagenet and \textit{tiered}Imagenet. The second column shows the number of parameters of each model in thousands (K). $^*Robust-20$ uses an 18-layer residual network. \texttt{--}Net is identical to EPNet but without EP. \textcolor{gray}{Gray colored results are obtained using $224\times 224$ pixels instead of the standard $84 \times 84$ pixel images.}}
\label{tab:main}
\begin{tabular}{@{}l|c|cc|cc@{}}
\toprule
 \multicolumn{1}{c}{} & \multicolumn{1}{c}{} & \multicolumn{2}{c}{\textbf{\textit{mini}Imagenet}} & \multicolumn{2}{c}{\textbf{\textit{tiered}Imagenet}}  \\
  \multicolumn{1}{c}{} & \multicolumn{1}{c}{Params} & \multicolumn{1}{c}{1-shot} & \multicolumn{1}{c}{5-shot} & \multicolumn{1}{c}{1-shot} & \multicolumn{1}{c}{5-shot} \\ \midrule
\multicolumn{6}{c}{CONV-4}\\
\midrule
Matching \cite{vinyals2016matching} & 112K & 43.56 \ci{0.84} & 55.31 \ci{0.73} & - & - \\
MAML \cite{liu2018learning} & 112K & 48.70 \ci{1.84} & 63.11 \ci{0.92} & 51.67 \ci{1.81} & 70.30 \ci{0.08}  \\
ProtoNet \cite{snell2017prototypical} & 112K & 49.42 \ci{0.78}  & 68.20 \ci{0.66}  & 53.31 \ci{0.89}  & 72.69 \ci{0.74}  \\ 
ReNet \cite{sung2018learning} & 223K & 50.44 \ci{0.82} & 65.32 \ci{0.70} & 54.48 \ci{0.92} & 71.32 \ci{0.78} \\ 
GNN \cite{garcia2017few} & 1619K & 50.33 \ci{0.36} & 66.41 \ci{0.63} & - & -  \\
TPN \cite{liu2018learning} & 171K & 53.75 \ci{0.86} & 69.43 \ci{0.67} & 57.53 \ci{0.96} & 72.85 \ci{0.74} \\ 
CC+rot \cite{gidaris2019boosting} & 112K & 54.83 \ci{0.43} & 71.86 \ci{0.33} & - & - \\\hline
EGNN \cite{kim2019edge} & 5068K & - & \multicolumn{1}{l}{$\;\;\textbf{76.37}$} & - & \multicolumn{1}{l}{$\;\;\textbf{80.15}$} \\ \hline
\texttt{--}Net (ours) & 112K & 57.18 \ci{0.83} & 72.57 \ci{0.66} & 57.60 \ci{0.93} & 73.30 \ci{0.74}  \\
EPNet (ours) & 112K & \textbf{59.32} \ci{0.88} & 72.95 \ci{0.64} & \textbf{59.97} \ci{0.95} & 73.91 \ci{0.75}  \\ 
\midrule
\multicolumn{6}{c}{RESNET-12}\\
\midrule
ProtoNets++ \cite{xing2019adaptive} & 7989K & 56.52 \ci{0.45} & 74.28 \ci{0.20} & 58.47 \ci{0.64} & 78.41 \ci{0.41}  \\
TADAM \cite{oreshkin2018tadam} & 7989K & 58.50 \ci{0.30} & 76.70 \ci{0.30} & - & - \\
MetaOpt-SVM~\cite{lee2019meta} & 12415K & 62.64 \ci{0.61} & 78.60 \ci{0.46} & 65.99 \ci{0.72} & 81.56 \ci{0.53} \\
TPN~\cite{liu2018learning} & 8284K & $\;\;59.46\;\;\;\;\;\;\;\;\;\;\;$ & $\;\;75.65\;\;\;\;\;\;\;\;\;\;\;$ & - & -  \\ 
$^*$Robust-20++ \cite{dvornik2019diversity} & 11174K & 58.11 \ci{0.64} & 75.24 \ci{0.49} & \textcolor{gray}{70.44 \ci{0.32}} & \textcolor{gray}{85.43 \ci{0.21}}\\
MTL~\cite{sun2019meta} & 8286K & 61.20 \ci{1.80} & 75.50 \ci{0.80} & - & - \\
CAN \cite{hou2019cross} & 8026K & 67.19 \ci{0.55} & 80.64 \ci{0.35} &  73.21 \ci{0.58} & 84.93 \ci{0.38} \\ 
\hline
\texttt{--}Net (ours) & 7989K & 65.66 \ci{0.85} & \textbf{81.28} \ci{0.62} & 72.60 \ci{0.91} & 85.69 \ci{0.65} \\ 
EPNet (ours) & 7989K & 66.50 \ci{0.89} & \textbf{81.06} \ci{0.60} & \textbf{76.53} \ci{0.87} & \textbf{87.32} \ci{0.64} \\
 \midrule
\multicolumn{6}{c}{WRN-28-10}\\
\midrule
LEO \cite{rusu2018meta} & 37582K & 61.76 \ci{0.08} & 77.59 \ci{0.12} & 66.33 \ci{0.05} & 81.44 \ci{0.09} \\
Robust-20++ \cite{dvornik2019diversity} & 37582K & 62.80 \ci{0.62} & 80.85 \ci{0.43} & - & -\\
wDAE-GNN \cite{gidaris2019generating} & 48855K & 62.96 \ci{0.15} & 78.85 \ci{0.10} & 68.18 \ci{0.16} & 83.09 \ci{0.12} \\
CC+rot \cite{gidaris2019boosting} & 37582K & 62.93 \ci{0.45} & 79.87 \ci{0.33} & 70.53 \ci{0.51} & 84.98 \ci{0.36} \\
Manifold mixup \cite{mangla2019charting} & 37582K & 64.93 \ci{0.48} & 83.18 \ci{0.72} & - & -  \\ \hline
\texttt{--}Net (ours) & 37582K & 65.98 \ci{0.85} & 82.22 \ci{0.66} & 74.04 \ci{0.93} & 86.03 \ci{0.63} \\
EPNet (ours) & 37582K & \textbf{70.74} \ci{0.85} & \textbf{84.34} \ci{0.53} & \textbf{78.50} \ci{0.91} & \textbf{88.36} \ci{0.57} \\

\bottomrule
\end{tabular}
\end{table}

\subsection{Implementation Details}
 For fair comparison with previous work, we used three common feature extractors: (i) a 4-layer convnet~\cite{vinyals2016matching,snell2017prototypical} with 64 channels per layer, (ii) a 12-layer resnet \cite{oreshkin2018tadam}, and (iii) a wide residual network (WRN-28-10) \cite{rusu2018meta,Zagoruyko2016WRN}. For \textit{mini}, \textit{tiered}Imagenet, and CUB, images are resized to $84 \times 84$. For Imagenet-FS, as described in \cite{gidaris2019boosting,gidaris2018dynamic} we use a resnet-18 and images are resized to $224\times224$.\smallskip
 
\noindent{\bf In the training stage}, the models are optimized using SGD with learning rate of $0.1$ for $100$ epochs. The learning rate is reduced by a factor of 10 every time the model reached a plateau, in which case the validation loss had not improved for 10 epochs. $\alpha$ is cross-validated on the 4-layer convnet.\smallskip

\noindent{\bf In the episodic fine-tuning stage} we randomly sample $5$ classes per episode, where in each class $k$ instances are selected for the support set and 15 for the query set.
Similar to the training stage, the model is optimized with SGD with learning rate $0.001$ reduced by a factor of 10 on plateau.  \smallskip

\noindent{\bf For training the wide residual networks (WRN)}, we apply the standard data augmentation methods mentioned by Szegedy \emph{et al.} ~\cite{szegedy2015going,rusu2018meta}. For the other architectures, we do not apply data augmentation.\smallskip

\noindent{\bf For Imagenet-FS}, we add EP right after the denoising autoencoder of wDAE-GNN~\cite{gidaris2019generating}. We use the original code provided by the authors.\footnote{\url{github.com/gidariss/wDAE\_GNN\_FewShot}}. Different from the other datasets, in this one evaluation is performed on the 311 test classes at the same time (\textit{311-way}), with the number of supports $k \in \{1, 2, 5, 10, 20\}$. \smallskip

We evaluate 3 variations of our method: (i) EPNet as described in Sec. \ref{sec:proposed_method}; (ii) \texttt{--}Net, which is identical to EPNet but without EP; and (iii) EPNet$_{SSL}$ (semi-supervised learning) as described in Sec. \ref{sec:method-semisup}. 

\subsection{Experimental Results}
In this section we compare EPNet with previous methods in the standard few-shot classification scenario, and semi-supervised learning scenarios. 

\begin{table}[t]
\centering
\setlength\tabcolsep{3pt}
\small
\caption{Comparison with the state of the art on CUB-200-2011. $^*Robust-20++$ uses an 18-layer residual network, and \textcolor{gray}{Accuracies obtained with $224\times 224$ images appear in gray.}}
\label{tab:main2}
\begin{tabular}{l|c|cc}
\toprule
  & backbone & 1-shot & 5-shot \\ \midrule
$^*$Robust-20++ \cite{dvornik2019diversity} & {\footnotesize RESNET-18} & \textcolor{gray}{68.68 \ci{0.69}} & \textcolor{gray}{83.21 \ci{0.44}}\\
EPNet (ours) & {\footnotesize RESNET-12} & \textbf{82.85} \ci{0.81} &\textbf{91.32} \ci{0.41} \\ \midrule
Manifold mixup \cite{mangla2019charting} & {\footnotesize WRN-28-10} & 80.68 \ci{0.81} & 90.85 \ci{0.44} \\
EPNet (ours) & {\footnotesize WRN-28-10} & \textbf{87.75} \ci{0.70} & \textbf{94.03} \ci{0.33} \\ \bottomrule
\end{tabular}
\end{table}
\begin{table}[t!]
\centering
\setlength\tabcolsep{3pt}
\small
\caption{Top-5 test accuracy on Imagenet-FS.}
\label{tab:imagenetfs}

\resizebox{\columnwidth}{!}{%
\begin{tabular}{l|cccccccccc}
\toprule
  & \multicolumn{5}{c}{Novel Classes} & \multicolumn{5}{c}{All classes} \\ 
  Approach & K=1 & 2 & 5 & 10& 20 & K=1 & 2 & 5 & 10 & 20 \\\midrule
Batch SGM  \cite{hariharan2017low} & - & - & - & - & - & 49.3 & 60.5 & 71.4 & 75.8 & 78.5\\
 PMN \cite{wang2018low}  & 45.8 & 57.8 & 69.0 & 74.3 & 77.4 & 57.6 & 64.7 & 71.9 & 75.2 & 77.5 \\ 
 LwoF \cite{gidaris2018dynamic} & 46.2 & 57.5 & 69.2 & 74.8 & 78.1 & 58.2 & 65.2 & 72.7 & 76.5 & 78.7 \\ 
 CC+ Rot \cite{gidaris2019boosting} & 46.43 \ci{ 0.24} & 57.80 \ci{ 0.16}& 69.67 \ci{ 0.09} &74.64 \ci{ 0.06} &77.31 \ci{ 0.05}& 57.88 \ci{ 0.15}& 64.76 \ci{ 0.10} & 72.29 \ci{ 0.07}& 75.63 \ci{ 0.04}& 77.40 \ci{ 0.03} \\
wDAE-GNN  \cite{gidaris2019generating} & 48.00 \ci{0.21} & 59.70 \ci{0.15} & 70.30 \ci{0.08} & 75.00 \ci{0.06} & 77.80 \ci{0.05} & 59.10 \ci{0.13} & 66.30  \ci{0.10} & 73.20 \ci{0.05} & 76.10 \ci{0.04} & 77.50 \ci{0.03} \\ \hline
wDAE-GNN + EP  (ours) & \textbf{50.07} \ci{0.27} &\textbf{62.16} \ci{0.16} &\textbf{72.89} \ci{0.11} &\textbf{77.25} \ci{0.07} &\textbf{79.48} \ci{0.05} &\textbf{60.87} \ci{0.16} &\textbf{68.53} \ci{0.10} &\textbf{75.56} \ci{0.07} &\textbf{78.28} \ci{0.04} &\textbf{78.89} \ci{0.03} \\ 
\bottomrule
\end{tabular}}
\end{table}

\begin{table}[t!]
\centering
\caption{SSL results with 100 unlabeled samples. \texttt{--}Net is identical to EPNet but without embedding propagation. *Re-implementation~\cite{yu2020transmatch}}
\label{tab:ssl_large}
\resizebox{\columnwidth}{!}{
\begin{tabular}{@{}l|c|cccc@{}}
\toprule
 &  & \multicolumn{2}{c}{\textbf{\textit{mini}Imagenet}}  &  \multicolumn{2}{c}{\textbf{\textit{tiered}Imagenet}} \\
  & Backbone & 1-shot & 5-shot  &  1-shot & 5-shot \\
 
 \midrule
TPN$_{SSL}$ \cite{liu2018learning} & CONV-4 & $\;\;52.78\;\;\;\;\;\;\;\;\;\;\;$ & $\;\;66.42\;\;\;\;\;\;\;\;\;\;\;$ &  $\;\;55.74\;\;\;\;\;\;\;\;\;\;\;$ & $\;\;71.01\;\;\;\;\;\;\;\;\;\;\;$ \\
k-Means$_{masked,soft}$ \cite{ren2018metalearning} & CONV-4 & 50.41 \ci{0.31} & 64.39 \ci{0.24} & - & - \\
EPNet (ours) & CONV-4 & 59.32 \ci{0.88} & 72.95 \ci{0.64} & 59.97 \ci{0.95} & 73.91 \ci{0.75} \\ 
\texttt{--}Net$_{SSL}$ (ours) & CONV-4 & 63.74 \ci{0.97} & 75.30 \ci{0.67} & 65.01 \ci{1.04} & 74.24 \ci{0.80}  \\
EPNet$_{SSL}$ (ours) & CONV-4 & \textbf{65.13} \ci{0.97} & \textbf{75.42} \ci{0.64} & \textbf{66.63} \ci{1.04} & \textbf{75.70} \ci{0.74} \\
\midrule
LST \cite{sun2019learning} & RESNET-12 & 70.10 \ci{1.90} & 78.70 \ci{0.80} & 77.70 \ci{1.60} & 85.20 \ci{0.80}  \\ 
EPNet (ours) & RESNET-12 & 66.50 \ci{0.89} & 81.06 \ci{0.60} & 76.53 \ci{0.87} & 87.32 \ci{0.64} \\
\texttt{--}Net$_{SSL}$ (ours) & RESNET-12 & 73.42 \ci{0.94} & 83.17 \ci{0.58} & 80.26 \ci{0.96} & 88.06 \ci{0.59} \\
EPNet$_{SSL}$ (ours) & RESNET-12 & \textbf{75.36} \ci{1.01} & \textbf{84.07} \ci{0.60} & \textbf{81.79} \ci{0.97} & \textbf{88.45} \ci{0.61}   \\ \midrule
*k-Means$_{masked,soft}$ \cite{ren2018metalearning} & WRN-28-10 & 52.78 \ci{0.27} & 66.42 \ci{0.21} & - & - \\
TransMatch \cite{yu2020transmatch} & WRN-28-10 & 63.02 \ci{1.07} & 81.19 \ci{0.59} & - & - \\
EPNet (ours) & WRN-28-10 & 70.74 \ci{0.85} & 84.34 \ci{0.53} & 78.50 \ci{0.91} & 88.36 \ci{0.57} \\
\texttt{--}Net$_{SSL}$ (ours) & WRN-28-10 & 77.70 \ci{0.96} & 86.30 \ci{0.50} & 82.03 \ci{1.03} & 88.20 \ci{0.61} \\
EPNet$_{SSL}$ (ours) & WRN-28-10 & \textbf{79.22} \ci{0.92} & \textbf{88.05} \ci{0.51} & \textbf{83.69} \ci{0.99} & \textbf{89.34} \ci{0.59} \\
\bottomrule
\end{tabular}}
\end{table}

\begin{table}[t!]
\centering
\caption{SSL results for the 5-shot 5-way scenario with different amounts of unlabeled data. The percentages refer to the amount of supports that are labeled in a set of 5 images per class.}
\label{tab:ssl}
\begin{tabular}{@{}l|c|ccccc@{}}
\toprule
 & Params & 20\% & 40\% & 60\% & 100\% &  \\ \midrule
GNN \cite{garcia2017few} & 112K & 52.45 & 58.76 & - & 66.41 &  \\
EGNN \cite{kim2019edge} & 5068K & \textbf{63.62} & 64.32 & 66.37 & \textbf{76.37} &  \\ \hline
\texttt{--}Net$_{SSL}$ (ours) & 112K & 58.52 \ci{0.97} & 64.46 \ci{0.79} & 67.81 \ci{0.74} &  57.18 \ci{0.83} \\
EPNet$_{SSL}$ (ours) & 112K & 60.66 \ci{0.97} & \textbf{67.08} \ci{0.80} & \textbf{68.74} \ci{0.74} & 59.32 \ci{0.88}  \\ \bottomrule
\end{tabular}
\end{table}

\paragraph{Main Results.} 
We first report the results for the methods that do not use an unlabeled set. As seen in Tables~\ref{tab:main} and \ref{tab:main2}, EPNet obtains state-of-the-art accuracies on \textit{mini}Imagenet, \textit{tiered}Imagenet, and CUB-200-2011 for the 1-shot and 5-shot benchmarks even when compared with models that use more parameters or higher resolution images. It can also be observed the effectiveness of EP in isolation when comparing EPNet with an identical model without EP (\texttt{--}Net). Higher-way and 10-shot results can be found in the supplementary material.
Note that EGNN \cite{kim2019edge} uses $\times45$ parameters. 
On the large-scale Imagenet-FS, EP improves all benchmarks by approximately $2\%$ accuracy, see Table \ref{tab:imagenetfs}. These results demonstrate the scalability of our method and the orthogonality with other embedding transformations such as denoising autoencoders~\cite{gidaris2019generating}. 

\paragraph{Semi-supervised learning.} We evaluate EPNet on the SSL setting where 100 additional unlabeled samples are available~\cite{ren2018metalearning,liu2018learning} (EPNet$_{SSL}$). We observe in Table \ref{tab:ssl_large} that including unlabeled samples increases the accuracy of EPNet for all settings, surpassing the state of the art by a wide margin of up to 16\% accuracy points for the 1-shot WRN-28-10. Similar to previous experiments, removing EP to EPNet (\texttt{--Net}) is detrimental for the model, supporting our hypotheses. Following the same setting described in \cite{garcia2017few,kim2019edge}, we trained our model in the 5-shot 5-way scenario where the support samples are partially labeled. In Table \ref{tab:ssl}, we report the test accuracy with \texttt{conv-4} when labeling 20\%, 40\%, 60\% and 100\% of the support set. EPNet obtains up to $2.7\%$ improvement over previous state-of-the-art when 40\% of the support are labeled. Moreover, EPNet also outperforms EGNN \cite{kim2019edge} in the 40\% and 60\% scenarios, although EPNet has $45\times$ less parameters. 

\subsection{Ablation Studies}
In this section we investigate the different hyperparameters of our method and the properties derived from embedding propagation. Additional ablations are provided in the supplementary material.

\begin{table}[t]
\centering
\caption{Algorithm ablation with conv-4 on 1-shot \textit{mini}Imagenet. EFT: Episodic Fine-tuning, ROT: Rotation loss, LP: Label Propagation, EP: Embedding Propagation }
\label{tab:pipeline-ablation}
\resizebox{\columnwidth}{!}{
\setlength{\tabcolsep}{0.2em}
\begin{tabular}{l|c|c|c|c|c|c|c|c|c|c|c|c|c|c|c|c}
\toprule
EXP &1 &2 &3 &4 &5 &6 &7 &8 &9 &10 &11 &12 &13 &14 &15 &16 \\ \hline
 EFT & & & & & & & & & \checkmark & \checkmark & \checkmark & \checkmark & \checkmark & \checkmark & \checkmark & \checkmark \\
 ROT &  &  &  &  & \checkmark & \checkmark & \checkmark & \checkmark &  &  &  &  & \checkmark & \checkmark & \checkmark & \checkmark \\
 LP &  &  & \checkmark & \checkmark &  &  & \checkmark & \checkmark &  &  & \checkmark & \checkmark &  &  & \checkmark & \checkmark \\
 EP &  & \checkmark &  & \checkmark &  & \checkmark &  & \checkmark &  & \checkmark &  & \checkmark &  & \checkmark &  & \checkmark \\ \hline
 ACC& 49.57& 52.83& 53.40& 55.75& 50.83& 53.63& 53.38& 55.55& 54.29& 56.38& 56.93& 58.35& 54.92& 56.46& 57.35& 58.85\\

\bottomrule
\end{tabular}}
\end{table}

\paragraph{Algorithm ablation.}  In Table \ref{tab:pipeline-ablation}, we investigate the impact of the rotation loss (ROT), embedding fine-tuning (EFT), label propagation (LP), and embedding propagation (EP) on the 1-shot \textit{mini}Imagenet accuracy. When label propagation is deactivated, we substitute it with a prototypical classifier. Interestingly, it can be seen that the improvement is larger when using LP in combination with EP (Table \ref{tab:pipeline-ablation}; columns 2-4, and 10-12). This finding is in accordance with the hypothesis that EP performs manifold smoothing, and this is beneficial for transductive and SSL algorithms. We included a rotation loss for fair comparison with other SotA \cite{gidaris2019boosting,mangla2019charting}, however, we see that the main improvement is due to the combination of EP with LP. We also find that episodic fine-tuning successfully adapts our model to the episodic scenario (Table \ref{tab:pipeline-ablation}; line 2). 

\begin{figure}[t!]
    \centering
    \includegraphics[width=\linewidth]{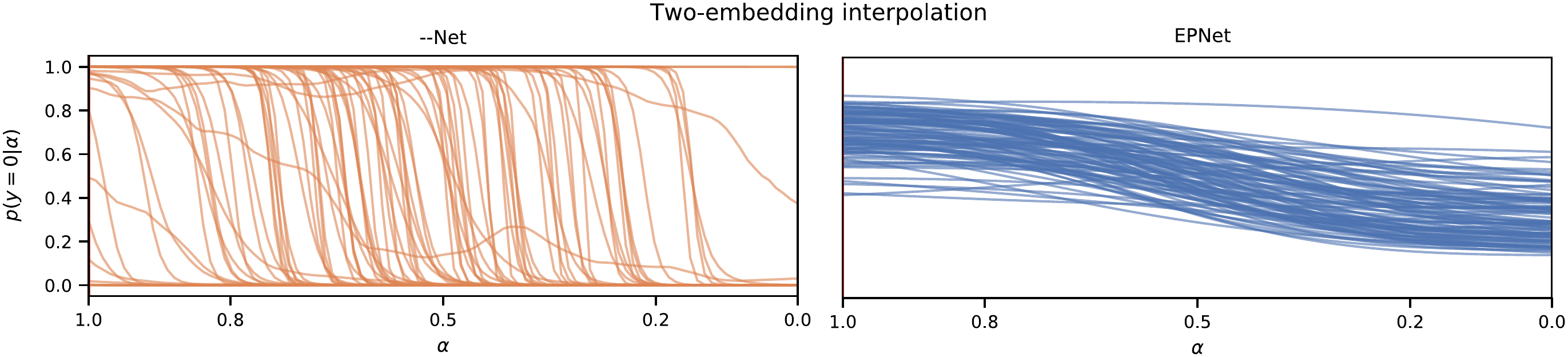}
    \caption{Interpolation of embedding pairs of different classes vs probability of belonging to the first of these two classes. The top row shows the class probability for resnet-12 embeddings extracted from EPNet, and the second(\texttt{--}Net) from the same network trained without embedding propagation. The scalar $\alpha$ controls the weight of the first embedding in the linear interpolation.}
     \label{fig:interpolation}
\end{figure}

\begin{figure}[t!]
    \centering
    \begin{subfigure}{0.45\textwidth}
    \centering
    \includegraphics[width=\textwidth]{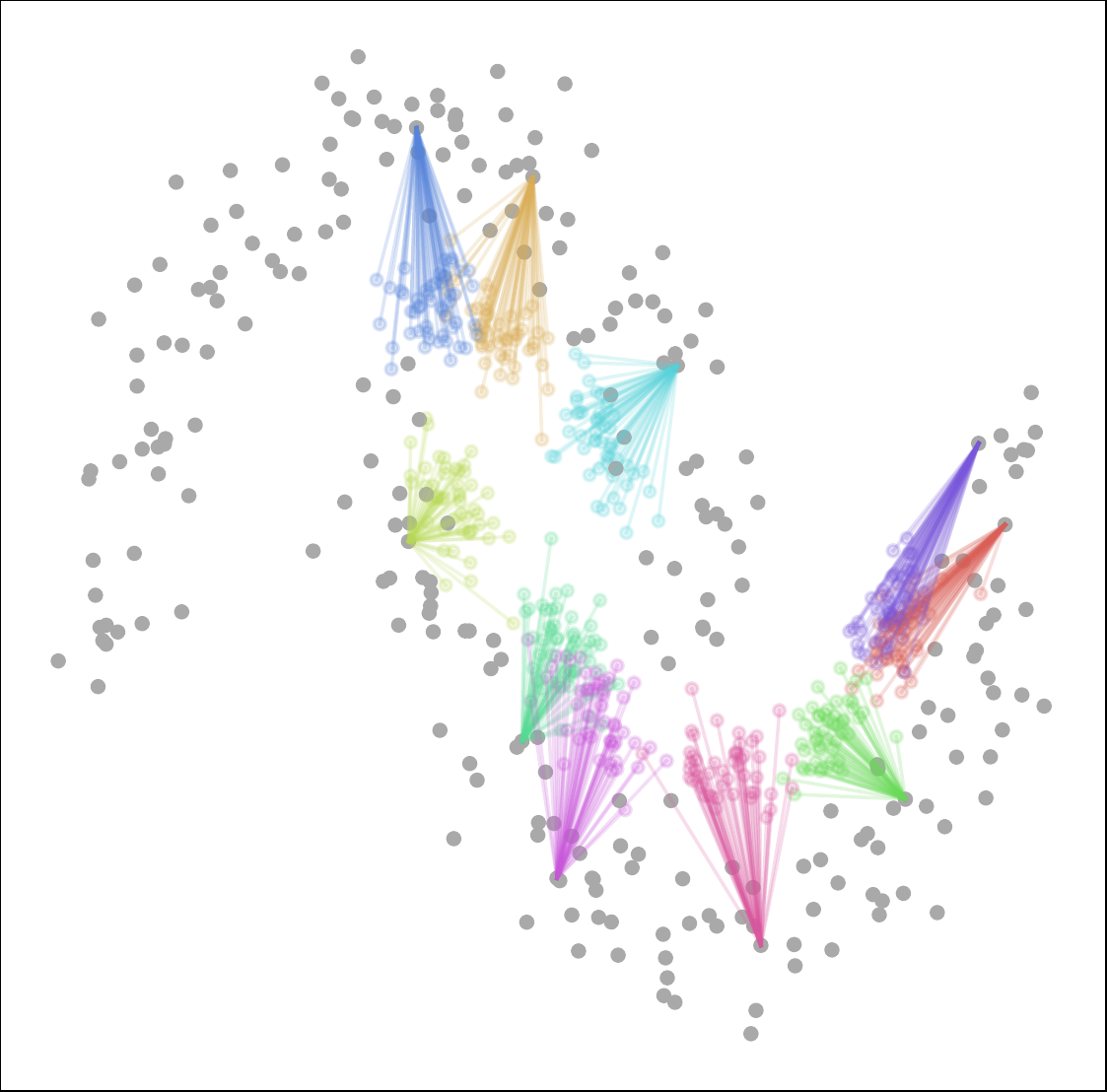}
    \caption{Pre-training phase} 
    \end{subfigure}
    \begin{subfigure}{0.45\textwidth}
    \centering
    \includegraphics[width=\textwidth]{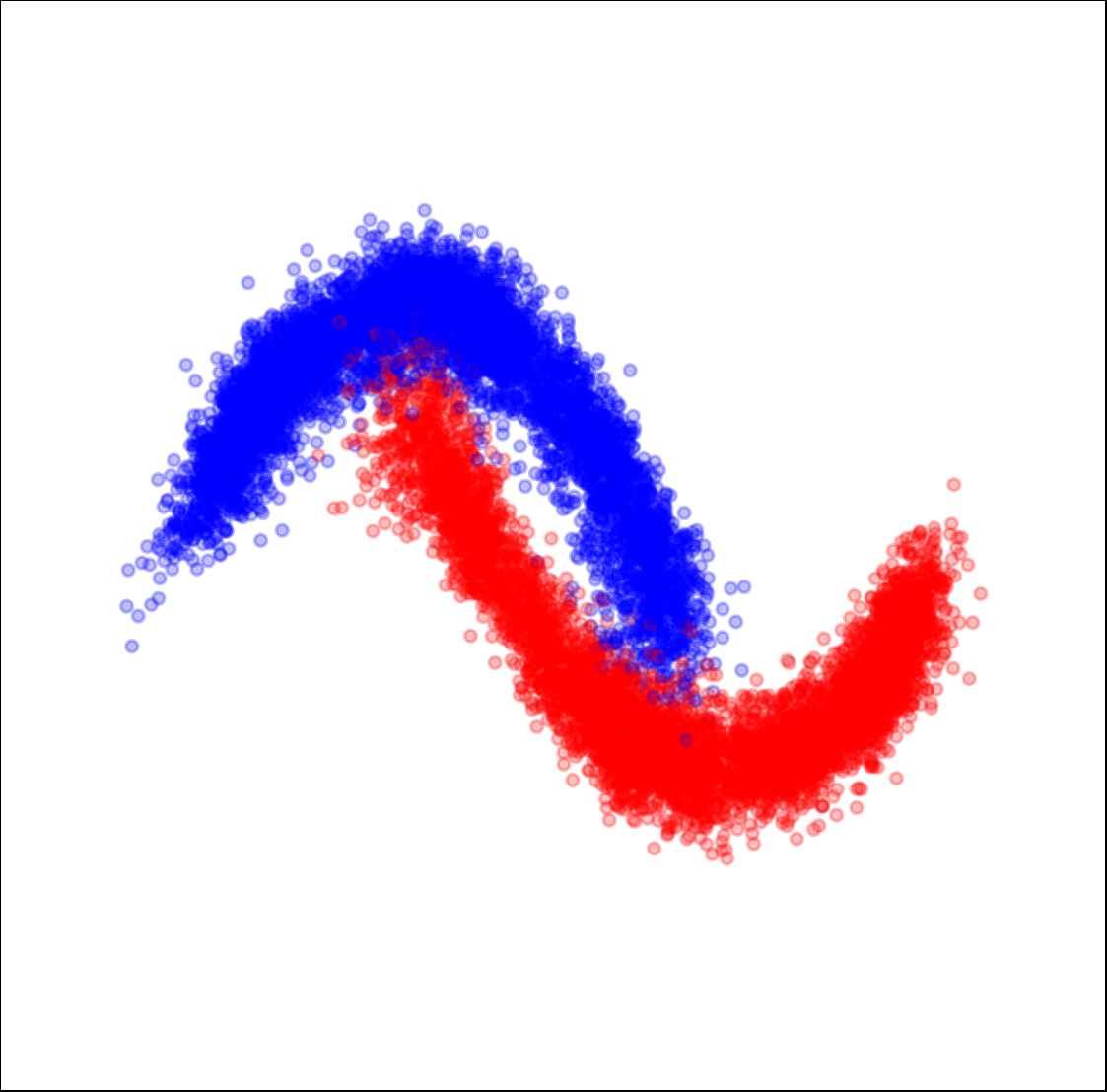}
    \caption{Propagated embeddings} 
    \end{subfigure}    
    \caption{Visualization of embedding propagation on the two moons dataset. \textit{The embeddings are shown on the same scale.}}
    \label{fig:my_label}
\end{figure}

\paragraph{Embedding propagation on manifold smoothness.} We explore whether  embedding propagation helps the classifier to attain smoother decision boundaries. We use EPNet to obtain image embeddings and select a set of random pairs $\feat_i, \feat_j$ that belong to different classes $y_i,y_j$. We then interpolate between each pair as $\pfeat = \alpha \cdot \feat_i + (1 - \alpha) \feat_j$ where $\alpha \in [0..1]$, and plot this value against $p(y_i | \pfeat)$ (Sec. \ref{eq:lp-class}) in Fig. \ref{fig:interpolation}. We also plot $p(y_i | \hat{\feat})$ where embeddings were obtained using EPNet without embedding propagation (\texttt{--}Net). We observe that EPNet has significantly smoother probability transitions than \texttt{--}Net as the embedding $\pfeat$ changes from $\feat_i$ to $\feat_j$. In contrast, \texttt{--}Net yields sudden probability transitions. This suggests that embedding propagation encourages smoother decision boundaries. 

In Figure~\ref{fig:my_label}, we show the effect of embedding propagation on a toy dataset. The dataset consists of embeddings that are arranged in two disjoint moons. The embeddings in the top moon belong to first class, and the other to the second class. Figure~\ref{fig:my_label}a) illustrates the effect of batch sampling during the pre-training phase. Gray points correspond to the extracted embeddings when no embedding propagation was applied. Each unique color shows multiple projections of the same gray point. Each projection is performed using a different batch of the data. This suggests that the projections of the same embedding fill a wide space in the manifold. As a result, the density and smoothness at the inter-class boundary increases. Figure~\ref{fig:my_label}b) shows the result of applying embedding propagation on all the gray points in Figure~\ref{fig:my_label}a). The blue and red colors correspond to the two-moon classes.  We observe that the propagated manifold is denser and more compact than the original one, possibly reducing the noise of the representations.

\section{Conclusion}
\label{sec:conclusion}

We have presented EPNet, an approach to address the problem of distribution shift few-shot learning. EPNet introduces a simple embedding propagation step to regularize the feature representations. Empirically, we have shown that embedding propagation smooths decision boundaries, which is an important factor for generalization. By leveraging the properties of smooth manifolds, we have shown significant improvements on the transductive, and semi-supervised learning setup compared to methods that do not use embedding propagation. As a result, EPNet achieves state-of-the-art results on \textit{mini}Imagenet, \textit{tiered}Imagenet, and CUB for the standard and semi-supervised scenarios. Further, we have shown that EP scales to the larger Imagenet-FS dataset, improving by more than 2\% the accuracy of the state-of-the-art wDAE-GNN \cite{gidaris2019generating} in all setups. We have compared EPNet with a non-smooth version of the same model (\texttt{--}Net), showing that smoothing alone accounts for $4.8\%$ accuracy improvement on 1-shot \emph{mini}Imagenet with Wide Residual Networks. 

\clearpage
%
%
\bibliographystyle{splncs04}
\bibliography{egbib}

\clearpage
\appendix
\section*{Appendix}
In the following sections, we provide results in the 10-shot scenario as well as more challenging settings, such as 10-way, 15-way, and 20-way classification. We also illustrate the effect of ablating different parts of the propagator matrix. Finally, we report the CO$_2$ emissions to produce this research.

\subsection*{10-shot test accuracy.} We report the results of EPNet and EPNet$_{SSL}$ in Table~\ref{tab:10shot}. Our methods improve the accuracy over \textit{TADAM} \cite{oreshkin2018tadam} and \textit{Discriminative} \cite{bauer2017discriminative} by $5\%$ accuracy. Since not many methods have been evaluated on this benchmark, it is challenging to illustrate the impact of embedding propagation. Further, we see that EPNet$_{SSL}$ does not improve much over EPNet, suggesting that embedding propagation has larger impact with fewer labeled data.

\begin{table}[h!]
\centering
\caption{10-shot test accuracy on \textit{mini}Imagenet, \textit{tiered}Imagenet, and CUB.}
\begin{tabular}{lccc}
\toprule
 & \textbf{mini} &\textbf{tiered} &\textbf{CUB} \\
  \midrule
\multicolumn{4}{c}{RESNET-12}\\\hline
Discriminative~\cite{bauer2017discriminative}&  78.50 & -     &-     \\
TADAM~\cite{oreshkin2018tadam}& 80.80 & -     &-     \\
EPNet (ours)& \textbf{85.39}  & \textbf{88.39}  &\textbf{92.68}      \\
EPNet$_{SSL}$ (ours) & \textbf{87.34}   &\textbf{89.24}    & \textbf{92.88 }     \\
\midrule
\multicolumn{4}{c}{WRN-28-10}\\
\hline
EPNet (ours)& \textbf{87.03}  & \textbf{89.46}     &\textbf{93.99}     \\
EPNet$_{SSL}$ (ours) & \textbf{89.02}    &\textbf{89.56}   & \textbf{94.11}\\\bottomrule
\label{tab:10shot}
\end{tabular}
\vspace{-1em}
\end{table}

\subsection*{Higher-way results}
For higher way setups, we compare against previous state-of-the-art methods on \textit{mini}Imagenet (Table~\ref{tab:n-way}). For a fair comparison, we use a WRN-28-10~\cite{Zagoruyko2016WRN} as our feature extractor. Our results show that EPNet attains higher test accuracies than previous state-of-the-art in all settings. For instance, in the 1-shot 20-way scenario, EPNet improves results from 36.5\% to 38.6\% accuracy. This suggests that embedding propagation generalizes effectively to higher number of classes. 

\begin{table}[h!]
\centering
\caption{\textit{mini}Imagenet 1-shot and 5-shot test accuracies for the 10-way, 15-way and 20-way scenarios. We report the accuracy with 95\% confidence intervals over 600 episodes. }
\label{tab:n-way}
\resizebox{\textwidth}{!}{\begin{tabular}{@{}lcccccc@{}}
\toprule
 & \multicolumn{2}{c}{\textbf{10-way}} & \multicolumn{2}{c}{\textbf{15-way}} & \multicolumn{2}{c}{\textbf{20-way}} \\
 & \textbf{1-shot} & \textbf{5-shot} & \textbf{1-shot} & \textbf{5-shot} & \textbf{1-shot} & \textbf{5-shot} \\ \midrule
Baseline++~\cite{chen2018a} & $40.43$ & $56.89$ & $31.96$ & $48.20$ & $26.92$ & $42.80$ \\
LEO~\cite{rusu2018meta} & $45.26$ & $64.36$ & $36.74$ & $56.26$ & $31.42$ & $50.48$ \\
DCO~\cite{lee2019meta} & $44.83$ & $64.49$ & $36.88$ & $57.04$ & $31.50$ & $51.25$ \\
Manifold mixup~\cite{mangla2019charting} & $50.40$ & $70.93$ & $41.65$ & $63.32$ & $36.50$ & $58.36$ \\
EPNet (ours) & $\textbf{53.70} \pm 0.59$ & $\textbf{72.17} \pm 0.44$ & $\textbf{44.55} \pm 0.28$ & $\textbf{64.44} \pm 0.34$ & $\textbf{38.55} \pm 0.19$ & $\textbf{59.01} \pm 0.27$ \\ \bottomrule
\end{tabular}}
\end{table}

\begin{table}[h!]
\centering
\caption{Propagator ablation with resnet-12 on 5-shot \textit{mini}Imagenet. Pre-training accuracy with (1) the full propagator matrix, (2) removing the diagonal of the propagator matrix, (3) removing the off-diagonal of the propagator matrix. As shown, our method leverages information from the neighborhood to attain optimal performance }
\label{tab:diag-ablation}
\begin{tabular}{c|c|c|c}
\toprule
\textbf{VER} & OFF-DIAG & DIAG & \textbf{ACC} \\
\hline
1 & $\checkmark$ & $\checkmark$ & 75.95 \ci{0.56} \\ \hline
2 & $\checkmark$ & - & 74.66 \ci{0.38} \\  \hline
3 & - & $\checkmark$ & 73.80 \ci{0.29} \\ \bottomrule
\end{tabular}
\end{table}
\subsection*{Additional ablation experiments} 
While in Table \ref{tab:pipeline-ablation} we ablate all the components of the proposed model, in Table \ref{tab:diag-ablation} we focus only on EP and whether it leverages information from neighboring embeddings (off-diagonal) or it just rescales them (diagonal). We show that neighbor information is important for the performance of EP. We train three versions of EPNet, (i) one with the full propagator matrix, (ii) one with only the off-diagonal of the matrix, and (iii) one with the diagonal matrix only. Hence, the second version only relies on information from neighboring embeddings to make predictions. The third version is equivalent to multiplying the original embeddings by a scalar. As seen in Table \ref{tab:diag-ablation}, the best performance is obtained with the first version, confirming that information from neighboring nodes is used. 

\subsection*{CO2 Emission Related to Experiments}

Experiments were conducted using a private infrastructure (located in Quebec, Canada) with a carbon emission factor of 0.02 kg/kWh. A cumulative of 24480 hours of computation was performed on hardware of type Tesla V100 (with a TDP of 250 W). Total emissions are estimated to be 146.88 kgCO$_2$eq, and 1000 kgCO$_2$eq (685\%) were offsetted through \href{https://www.goldstandard.org/take-action/offset-your-emissions}{Gold Standard}. Estimations were obtained using the \href{https://mlco2.github.io/impact#compute}{MachineLearning Impact calculator} \cite{lacoste2019quantifying}.

\end{document}